\documentclass[twoside,11pt,preprint]{article}

\usepackage{blindtext}

\usepackage[abbrvbib]{jmlr2e}

\usepackage{listings}
\usepackage[dvipsnames]{xcolor}
\usepackage{multirow}
\usepackage{booktabs}
\usepackage{tabularx}
\usepackage{makecell}
\usepackage{xcolor}
\usepackage[textsize=tiny]{todonotes}
\newcommand{\ccite}[1]{\tiny(\citetalias{#1})}

\definecolor{strings}{rgb}{.624,.251,.259}
\definecolor{keywords}{rgb}{.224,.451,.686}
\definecolor{comment}{rgb}{.322,.451,.322}

\lstdefinelanguage{python}{
  morekeywords={from, import, as, for, in, while, def, return, 
  =, +, -, /, *, lambda},
  keywords=[3]{ContinuousTimeStateEvolution, DiscreteTimeStateEvolution, DynamicalModel, FilterBasedMarginalLogLikelihood, Context, Trajectory},
  morecomment=[l]{\#},
  morecomment=[s]{"""}{"""},
  morestring=[b]',
  morestring=[b]",
  alsoletter={<>=-+/*},
  sensitive=true
}

\usepackage[most]{tcolorbox}
\newtcolorbox{docode}[1][]{
    colback=white,        %
    colframe=black,         %
    width=0.9\textwidth,    %
    arc=4mm,                %
    boxrule=0.5pt,          %
    left=2em,               %
    enhanced,
    fontupper=\ttfamily,    %
    #1                      %
}

\tcbuselibrary{listings}

\newtcblisting{boxedlisting}{
  listing only,
  colback=white,
  colframe=black,
  boxrule=0.5pt,
  arc=4pt,
  width=\textwidth,
  left,
  left=6pt,
  right=6pt,
  top=2pt,
  bottom=2pt,
  listing options={
    language=python,
    basicstyle=\fontsize{8pt}{8.25pt}\selectfont\ttfamily,
    keywordstyle=\color{BrickRed}\bfseries,
    commentstyle=\color{comment},
    stringstyle=\color{strings},
    showstringspaces=false,
    keepspaces=true,
    tabsize=2,
  }
}
\renewcommand{\texttt}[1]{\lstinline[basicstyle=\fontsize{11pt}{11.25pt}\selectfont\ttfamily]{#1}}

\usepackage{xcolor}

\newcommand{\given}{\,|\,}

\defcitealias{BarberChiappa2006VBLGSSM}{BC06}

\defcitealias{Aicher2019SGMCMCSSM}{AMFF19}
\defcitealias{NinnessHenriksen2010BayesianSID}{NH10}
\defcitealias{HelskeVihola2021bssm}{HV21}

\defcitealias{Linden2022UKFMCMC}{LKR22}
\defcitealias{erazo2017ukfmcmc}{EN17}
\defcitealias{Drovandi2022EnsembleMCMC}{DEGP22}

\defcitealias{Frigola2014VGPSSM}{FCR14}
\defcitealias{Naesseth2018VSMC}{NLRB18}

\defcitealias{Dahlin2015PMHGrad}{DLS15}
\defcitealias{nemeth2016particle}{NSF16}
\defcitealias{amri2025particle}{AEW25}

\defcitealias{aicher2025stochastic}{APNFF25}
\defcitealias{andrieu2010particle}{ADH10}

\defcitealias{Foti2014SVIHMM}{FXLF14}
\defcitealias{Ma2017SGMCMCHMM}{MFF17}
\defcitealias{Scott2002HMMBayes}{S02}

\defcitealias{Mbalawata2013SDEMCMC}{MSH13}
\defcitealias{Sarkka2015SDEAdaptiveMCMC}{SHMH15}
\defcitealias{GolightlyWilkinson2011BiochemPMCMC}{GW11}

\defcitealias{tran2017variational}{TNK17}
\defcitealias{stan_gaussian_dlm}{SDT24}

\defcitealias{sarkka2023bayesian}{SS23}

\defcitealias{ansari2023neural}{AHLS23}

\usepackage{lastpage}
\jmlrheading{23}{2026}{1-\pageref{LastPage}}{1/21; Revised 5/22}{9/22}{21-0000}{Daniel Waxman, Dmitry Batenkov, John Feser, Andy Zane, Eli Bingham, Youssef Marzouk, and Matthew E. Levine}

\ShortHeadings{Dynestyx: PPL For Dynamical Systems}{Waxman, Batenkov, Feser, Zane, Bingham, Marzouk, Levine}
\firstpageno{1}

\begin{document}

\title{Dynestyx: A Probabilistic Programming Library for Dynamical Systems}

\author{\name Daniel Waxman\textsuperscript{1,2,\ensuremath{\dagger}}
\email dan@basis.ai \\
\name Dmitry Batenkov\textsuperscript{1}
\email dima@basis.ai \\
\name John Feser\textsuperscript{1}
\email jack@basis.ai \\
\name Andy Zane\textsuperscript{1}
\email andy@basis.ai \\
\name Eli Bingham\textsuperscript{1}
\email eli@basis.ai \\
\name Youssef Marzouk\textsuperscript{2}
\email ymarz@mit.edu \\
\name Matthew E.\ Levine\textsuperscript{1,\ensuremath{\dagger}}
\email matt@basis.ai \\
\addr \textsuperscript{1}Basis Research Institute \textsuperscript{2}Massachusetts Institute of Technology
}

\editor{My editor}

\maketitle

\begin{abstract}%
State-space models (SSMs) are the standard formalism for Bayesian treatment of dynamical systems, with natural applications in statistics, signal processing, and machine learning. Despite their importance in both theory and application, dynamical systems have proven difficult to incorporate in modern probabilistic programming languages (PPLs), making state-of-the-art methods less accessible to practitioners and introducing friction in following the ``Bayesian workflow.''
 We introduce \texttt{dynestyx}, a probabilistic programming library with first-class support for SSMs, including state-of-the-art methods in the estimation of both states and parameters. Through a single, unified interface, users may specify arbitrary priors for discrete-time or continuous-time dynamical systems, perform inference over mixed-effect data, and make state and parameter estimates with principled uncertainty quantification.
\end{abstract}

\begin{keywords}
  state-space models, dynamical systems, probabilistic programming
\end{keywords}

\begin{table}[b!]
\raggedright
\footnoterule
\footnotesize\textsuperscript{\ensuremath{\dagger}}Corresponding authors.
\\
Website: \url{https://basisresearch.github.io/dynestyx}.
\\
Code: \url{https://github.com/BasisResearch/dynestyx}.
\end{table}

\section{Introduction}

\emph{Dynamical systems} are a modeling paradigm common to many applications of Bayesian machine learning, including neuroscience \citep{paninski2010new} and statistical signal processing \citep{sarkka2023bayesian}. A principal advantage of the dynamical modeling paradigm is the ability to directly model time evolution, feedback, and interaction---all fundamental to mechanistic understanding of the target phenomena. SSMs also provide interesting settings for theoretical development by posing a rich, recursive structure over correlated observations.
SSMs have therefore attracted much methodological attention, creating a vast literature on inference algorithms that exploit their structure. Unfortunately, these inference methods are often contained only in bespoke codebases and are thus under-utilized in many applications.

We introduce \texttt{dynestyx}
, an extension of the \emph{probabilistic programming language (PPL)} \texttt{NumPyro} \citep{phan2019composable} designed for inference in dynamical systems. \texttt{dynestyx} respects the PPL principle of separation of concerns (cf., \citealp[Sec.~2]{bingham2019pyro}) by providing a uniform interface for specifying dynamical models and making structure-exploiting Bayesian inference methods interchangeable: a single model specification can be paired with different approximations and algorithms and then composed directly with standard \texttt{NumPyro} inference workflows for rapid development and systematic method comparison. %

\section{Mathematical Description}

\paragraph{Defining SSMs} The formalism of dynamical systems that we adopt in \texttt{dynestyx} is that of state space models, where a latent, Markovian \emph{state} $x_t \in \mathbb{R}^{d_x}$ has an 
uncertain \emph{initial condition} with probability density $p(x_0)$
that is propagated through time; and inference proceeds through \emph{observations} $y_t \in \mathbb{R}^{d_y}$ that are coupled to the state. We allow both the evolution of the state $x_t$ and the observation process $y_t \given x_t$ to be mediated by a known, exogenous variable $u_t$, commonly known as a \emph{control}. The state $x_t$ may evolve in discrete time, i.e.,
\begin{equation} \label{eq:discrete} 
    x_{t+1} \sim p(x_{t+1} \given x_{t}, u_t, t),
\end{equation}
or in continuous time, i.e., from a stochastic differential equation (SDE),
\begin{equation} \label{eq:sde}
    \mathrm{d}x_t = f(x_t, u_t, t) \, \mathrm{d}t + L(x_t, u_t, t) \,  \mathrm{d}\beta_t, 
\end{equation}
where $f(x, u, t) \colon \mathbb{R}^{d_x} \times \mathbb{R}^{d_u} \times [0,T] \to \mathbb{R}^{d_x}$ 
is known as the \emph{drift term}, and $L(x, u, t) \,  \mathrm{d}\beta_t$ the \emph{diffusion term}, where $L \colon \mathbb{R}^{d_x} \times \mathbb{R}^{d_u} \times [0,T] \to \mathbb{R}^{d_x \times d_B}$ and $\beta_t$ is $d_B$-dimensional Brownian motion. In either case, inference must proceed via discrete-time observations,
\begin{equation} \label{eq:observations}
    y_t \sim p(y_t \given x_t, u_t, t).
\end{equation}

In many applications, we do not know the components of \eqref{eq:discrete}, \eqref{eq:sde}, or \eqref{eq:observations} \emph{a priori}, and instead specify parametric forms for each component. We say that latent dynamics for $x_t$ (via \eqref{eq:discrete} or \eqref{eq:sde}) are parameterized by $\theta \in \mathbb{R}^{d_\theta}$, and that the observation process $p(y_t \given x_t, u_t)$ is parameterized by 
$\varphi \in \mathbb{R}^{d_\varphi}$. Ensuing statistical questions involve pairings of $\left(\{x_t\}_{t=0}^T, \theta, \varphi\right)$.

\paragraph{Defining Posteriors for SSM} An SSM induces several objects of statistical interest; the \emph{filtering distribution} $p(x_t \given y_{1:t})$ encodes belief about the latent state $x_t$ after observing $y_{1:t}=\{y_1,\dots,y_t\}$. 
The related \emph{smoothing} distribution encodes posterior belief after observing future observations, i.e., it captures $p(x_\tau \given y_{1:t})$ for $\tau \leq t$.
Filtering and smoothing are intractable outside special cases, but admit many effective approximations (cf., \citealp{sarkka2023bayesian,law2015data}). 
Another central task is \emph{system identification}, i.e., inferring $\{\theta,\varphi\}$ in \eqref{eq:discrete}, \eqref{eq:sde}, and \eqref{eq:observations}. This proceeds via the marginal likelihood (ML) $p(y_{1:t}\given \theta,\varphi)$: typically intractable, but often well-approximated by filtering, with competing estimators. For instance, sequential Monte Carlo \citep{chopin2020introduction} yields unbiased but noisy and expensive estimates of the ML, while the ensemble Kalman filter \citep{evensen1994sequential} is efficient but biased. These likelihood estimates, along with a prior $p(\theta,\varphi)$,
then feed a parameter inference scheme (e.g., MCMC or variational inference), with its own tradeoffs.

\section{The Computational Frontier and \texttt{dynestyx}}
The state-of-the-art in Bayesian system identification primarily comprises two steps: (i) computing (an approximation of) the ML 
$\hat{\mathcal{Z}}_{1:T}(\theta, \varphi) \approx \mathcal{Z}_{1:T}(\theta, \varphi) \triangleq p(y_{1:T} \given \theta, \varphi)$, and (ii) using this information in some general parameter identification algorithm, using the log-target (i.e., the log-marginal posterior density)
$\log p(\theta, \varphi) + \log \hat{\mathcal{Z}}_{1:T}(\theta, \varphi)$.

A practical difficulty in performing the two-step system identification procedure above is that different filters and parameter inference methods have historically lived across distinct, bespoke codebases. This two-part procedure, however, suggests an opportunity for orthogonalization: if we can specify a model, ML estimator, and parameter inference algorithm separately, a large design space of algorithms are available by construction. This orthogonalization aligns with \texttt{dynestyx}'s respect for PPL separation of concerns.

In \texttt{dynestyx}, a model is specified as a Python function that (i) specifies the joint prior 
$p(\theta, \varphi)$ via an arbitrary \texttt{NumPyro} program, (ii) packages these to a \texttt{DynamicalModel}, and (iii) samples them with \texttt{dynestyx.sample(...)}. The \texttt{dynestyx.sample(...)} function has arguments (a) site name; (b) times and values of observations;
(c) times and values of controls;
and (d) times at which to return samples of the predictive distribution of the state and observations:
\vspace{-0.2cm}
\begin{boxedlisting}
def model(obs_times, obs_values, ctrl_times, ctrl_values, predict_times):
    theta = numpyro.sample("theta", p_theta)

    continuous_state_evolution = ContinuousTimeStateEvolution(
        drift=f_drift(theta),
        diffusion=ScalarDiffusion(0.1), # More efficient, but arbitrary JAX functions are allowed
    )
    
    dynamics = DynamicalModel(
        initial_condition=dist.Normal(0, 1),
        state_evolution=continuous_state_evolution, # or discrete_state_evolution
        observation_model=dist.Poisson(...),
    )

    return dsx.sample("f", dynamics, obs_times=obs_times, obs_values=obs_values,
                ctrl_times=ctrl_times, ctrl_values=ctrl_values, predict_times=predict_times)
\end{boxedlisting}%
We emphasize that this model \emph{does not} specify any particular inference method (i.e., filter or smoother) or numerical discretizer, as this would violate the PPL doctrine of separation of concerns. The model is thus a specification of the idealized data-generating process only, with any approximations or computation coming strictly at inference time.

At inference time, \texttt{dynestyx} computes approximate ML according to a user-specified interpretation of the \texttt{dsx.sample} statement (e.g., a particular filtering algorithm) via effect handlers~\citep{plotkin2009handlers,phan2019composable}.
The resulting model is compatible with arbitrary \texttt{NumPyro} inference routines. For example, an implementation of particle HMC \citep{amri2025particle} comprises standard \texttt{NumPyro} HMC code, under an appropriate \texttt{Filter} interpretation of \texttt{dsx.sample}; this \texttt{Filter} interpretation can be further passed to a \texttt{Simulator(...)} interpretation, which samples from the filtering posterior predictive:
\begin{boxedlisting}
from numpyro.infer import MCMC, HMC

with Simulator(): # Sample from the filtering posterior predictive at predict_times
    with Filter(ParticleFilterConfig(n_particles=1_000)): # Compute the filtering distribution
        with Discretizer(): # Euler-Maruyama Discretization
            mcmc = MCMC(HMC(model), num_samples=100, num_warmup=100)
            mcmc.run(jr.PRNGKey(0), obs_times=..., obs_values=..., 
                    ctrl_times=..., ctrl_values=..., predict_times=...)
            posterior_samples = mcmc.get_samples()
\end{boxedlisting}

\section{A Summary of Implemented Methods}
\texttt{dynestyx} implements a number of state estimation (i.e., filtering and smoothing) algorithms via integrations with existing software, including \texttt{dynamax} \citep{linderman2025dynamax}, \texttt{cd-dynamax} \citep{cd_dynamax}, and \texttt{cuthbert} \citep{Duffield_cuthbert_2026}. This results in an expressive landscape of possible methods, including Kalman filtering/RTS smoothing, extended Kalman filtering/RTS smoothing, unscented Kalman filtering, ensemble Kalman filtering, and (differentiable) particle filtering/smoothing.

Since \texttt{dynestyx} is designed to work with arbitrary \texttt{NumPyro} code, its models can be quite general, making resulting inference compatible with generic \texttt{JAX} routines, including variational inference, HMC/NUTS, stochastic gradient MCMC, gradient-informed MCMC like MALA, and generic MH-type algorithms. Each of these methods presents its own tradeoffs (e.g., speed or bias) and interacts uniquely with the corresponding SSM.

Many combinations of a ML estimator and a Bayesian parameter inference method map cleanly into previous efforts in the literature, while others, to the best of our knowledge, were first implemented here. We summarize implemented methods, and corresponding references, in Table~\ref{tab:lit_review}. Methods that have not previously appeared in the literature (at least explicitly), such as (CD) EnKF with gradient-informed MCMC, are marked with ``+''.

\texttt{dynestyx} additionally implements mixed-effect dynamical systems models \citep{wang2014estimating, picchini2010stochastic} via a \texttt{plate} primitive. These mixed-effect models are compatible with any combination of ML estimation/parameter inference method as in Table \ref{tab:lit_review}.

\begin{table*}[t]
\centering
\scriptsize
\setlength{\tabcolsep}{4pt}
\renewcommand{\arraystretch}{1.15}

\begin{tabular}{lcccc}
\hline
& Variational Bayes & Gradient-Informed MCMC & SG-MCMC & Generic MH\\
\hline

KF
& \ccite{BarberChiappa2006VBLGSSM}
& \ccite{stan_gaussian_dlm}
& \ccite{Aicher2019SGMCMCSSM}
& \ccite{NinnessHenriksen2010BayesianSID} \\

EKF
& +
& \ccite{sarkka2023bayesian}
& +
& \ccite{HelskeVihola2021bssm} \\

UKF
& +
& \ccite{sarkka2023bayesian}
& +
& \ccite{Linden2022UKFMCMC}, \ccite{erazo2017ukfmcmc} \\

EnKF
& +
& +
& +
& \ccite{Drovandi2022EnsembleMCMC} \\

PF
& \ccite{Frigola2014VGPSSM}, \ccite{tran2017variational} \ccite{Naesseth2018VSMC}
& \ccite{Dahlin2015PMHGrad}, \ccite{nemeth2016particle}, \ccite{amri2025particle}
& \ccite{aicher2025stochastic}
& \ccite{andrieu2010particle} \\

HMM
& \ccite{Foti2014SVIHMM}
& +
& \ccite{Ma2017SGMCMCHMM}, \ccite{Aicher2019SGMCMCSSM}
& \ccite{Scott2002HMMBayes} \\

CD KF
& +
& \ccite{Mbalawata2013SDEMCMC}
& +
& \ccite{Mbalawata2013SDEMCMC} \\

CD EKF
& \ccite{ansari2023neural}
& \ccite{Mbalawata2013SDEMCMC}
& +
& \ccite{Sarkka2015SDEAdaptiveMCMC}, \ccite{HelskeVihola2021bssm} \\

CD UKF
& +
& +
& +
& \ccite{Sarkka2015SDEAdaptiveMCMC}, \ccite{Linden2022UKFMCMC} \\

CD EnKF
& +
& +
& +
& + \\

CD PF
& +
& +
& +
& \ccite{GolightlyWilkinson2011BiochemPMCMC}, \ccite{Sarkka2015SDEAdaptiveMCMC} \\

\hline
\end{tabular}
\caption{Mapping combinations of ML estimators (e.g., KF, EKF, etc.) and Bayesian inference methods (variational Bayes, HMC/NUTS, etc.) 
to existing literature. The prefix ``CD'' indicates a paper explicitly considers continuous-discrete SSMs (i.e., SSMs specified by \eqref{eq:sde} and \eqref{eq:observations}). Entries with ``+'' denote combinations available in \texttt{dynestyx} that do not appear in the literature, to the best of our knowledge.}\label{tab:lit_review}
\end{table*}

\section{Conclusions} Unified representation and inference for SSMs is powerful: we can directly compare different compositions of inference methods and establish rules of thumb and pathologies that occur in application of complex, flexible, and potentially mis-specified models. This further invites new innovations in methodology and creates a natural testbed for those advancements.

\acks{We thank Tim Cooijmans, Sam Duffield, Markus Heinonen, Martin Lysy, Dat Nyugen, Theo Rashid, Rafal Urbaniak, Iñigo Urteaga, and Yixiu Zhao for useful discussions. We are grateful for funding support to Basis Research Institute from private donations.
}

\bibliography{main}

\end{document}